\documentclass[lettersize,journal]{IEEEtran}
\usepackage{amsmath,amsfonts}
\usepackage{algorithmic}
\usepackage{algorithm}
\usepackage{array}
\usepackage[caption=false,font=normalsize,labelfont=sf,textfont=sf]{subfig}
\usepackage{textcomp}
\usepackage{stfloats}
\usepackage{url}
\usepackage{verbatim}
\usepackage{graphicx}
\usepackage{cite}
\usepackage{multirow}
\usepackage{amssymb}
\usepackage{booktabs}
\usepackage{float}
\usepackage{hyperref}
\hypersetup{
    colorlinks=true,
    linkcolor=blue,
    filecolor=blue,      
    urlcolor=blue,
    citecolor=cyan,
}
\begin{document}

\title{EGVD: Event-Guided Video Deraining}

\author{Yueyi Zhang, Jin Wang, Wenming Weng, Xiaoyan Sun, Zhiwei Xiong
\thanks{The authors are with the Department of Electronic Engineering and Information Science, University of Science and Technology of China, Hefei, 230026, China. E-mail: zhyuey@ustc.edu.cn; jin01wang@mail.ustc.edu.cn; wmweng@mail.ustc.edu.cn; sunxiaoyan@ustc.edu.cn; zwxiong@ustc.edu.cn}}



\maketitle

\begin{abstract}
With the rapid development of deep learning, 
video deraining has experienced significant progress. 
However, existing video deraining pipelines cannot achieve satisfying performance for scenes with rain layers of complex spatio-temporal distribution. In this paper, we approach video deraining by employing an event camera. 
As a neuromorphic sensor, the event camera suits scenes of non-uniform motion and dynamic light conditions. 
We propose an end-to-end learning-based network to unlock the potential of the event camera for video deraining. 
First, we devise an event-aware motion detection module to adaptively aggregate multi-frame motion contexts using event-aware masks. 
Second, we design a pyramidal adaptive selection module for reliably separating the background and rain layers by incorporating multi-modal contextualized priors. 
In addition, we build a real-world dataset consisting of rainy videos and temporally synchronized event streams. 
We compare our method with extensive state-of-the-art methods on synthetic and self-collected real-world datasets, demonstrating the clear superiority of our method. The code and dataset are available at   \url{https://github.com/booker-max/EGVD}.
\end{abstract}


\section{Introduction}
\IEEEPARstart{O}{utdoor} cameras encounter adverse weather conditions, such as rain. Rain not only degrades the visual quality of captured images and videos but also hampers the performance of downstream multimedia tasks that rely on clean video frames, such as object tracking \cite{nai2023robust,wang2023spatio}, person re-identification (Re-ID) \cite{ sun2022multi,wang2023weakly, bai2021sanet} and SLAM \cite{bing2022toward, zhao2021probabilistic, shao2020tightly}. Under the considerable demands of rain-free videos, it is imperative to explore the algorithm of video deraining.

Recently, many methods are proposed to handle the video deraining task and obtain substantial performance on some public benchmark datasets, e.g., NTURain \cite{chen2018robust}, RainSynLight25 and RainSynComplex25 \cite{liu2018erase}. 
However, there still exist some drawbacks. 
On one hand, most of the methods are insufficient to model the spatio-temporal distribution of rain layers that exhibit strong spatial variations e.g., scale, direction, and density) and temporal dynamics (e.g., velocity and acceleration).
Many deraining methods fail in accurately modeling these randomly scattered rain streaks, resulting in unsatisfactory rain streak removal and detail loss in non-rain regions. 
On the other hand, for exploiting multi-frame correlation, existing video deraining methods \cite{yang2019frame, yang2020self, zhong2021star, mu2021triple} follow the flow-based pipeline.
They utilize either optical flow or deformable convolution \cite{yan2021self} to temporally align neighboring frames for rain removal.
However, the presence of rain streaks breaks the existing flow constraints (the brightness constancy constraint) and prohibits estimating an accurate motion field for alignment, especially under torrential rainfall. Hence, these methods cannot make full use of information from neighboring frames, calling for more effective solutions.
To sum up, how to precisely model the spatio-temporal distribution of rain layers blended in the rainy video and how to learn the favorable clues from the adjacent frames are worth more attention. 

Instead of struggling to design complex computational architectures, we resort to using an event camera \cite{gallego2020event}, an emerging bio-inspired sensor, to solve the aforementioned limitations of existing video deraining methods.
Event cameras are novel vision sensors whose working mechanism is drastically different from conventional frame sensors. 
Instead of capturing images at regular intervals, they report pixel-wise changes in brightness as a stream of asynchronous events.
With unique advantages such as high temporal resolution (up to 1 MHz), high dynamic range (up to 140 dB), and very low power consumption, event cameras have already been applied in a variety of video tasks.

Actually, the moving rain streaks usually produce obvious intensity changes, which naturally suits the dynamic perception of event cameras. We explore the application of event cameras in the context of video deraining from two perspectives.
First, we use an event camera as a complementary sensor, which provides additional motion-aware information that is not explicitly provided by conventional frame cameras.
With this hybrid imaging system, not only can we acquire absolute pixel intensity measurements reflecting rain and background layers, but also the motions of rain and moving background objects are prominently detected by the event camera.
Second, it is of great significance to separate rain and background information that is typically merged in the image and feature domain. Nevertheless, it is hard to be achieved by conventional frame cameras especially when rain layers are complex because the motion prior out of exposure time is inaccessible.
In comparison, event cameras, which output data at microseconds, can accurately perceive the motion variation of background layers.
Therefore, the visibility of rain and moving objects is enhanced by event cameras and served as strong guidance for subsequent rain removal.

In this paper, we propose an end-to-end learning neural network, called \textbf{E}vent-\textbf{G}uided \textbf{V}ideo \textbf{D}eraining Network (\textbf{EGVD}), for video deraining with an event camera. 
To be concrete, we first devise an event-aware motion detection module to selectively detect and aggregate motion information of neighboring frames using event-aware masks. Thus, we acquire fused frame features containing rain-background motions enhanced by event streams.
Second, in contrast to prior works that only exploit one modality (i.e., frame), we design a pyramidal adaptive selection module to reliably separate rain and background layers in the feature domain via incorporating multi-modal contextualized clues.
In such a way, we reconstruct the final rain layer, which is then added to the rain-degraded input to produce a final rain-free video.

For training our network, we generate large-scale synthetic datasets, including various rains from light drizzling to heavy falls. For real-world evaluation, we use a Color-DAVIS346 camera \cite{taverni2018front} to build a real-world dataset for event-based video deraining, which contains rain-degraded videos and temporally synchronized events.

The main contributions are summarized as follows:
\begin{itemize}
\item We approach video deraining with an event camera by exploiting its motion-aware imaging and high temporal resolution property.
\item We design two novel components, i.e., an event-aware motion detection module and a pyramidal adaptive selection module, for effectively enhancing motion-aware regions and separating rain-background layers to produce rain-free videos.
\item We build a real-world dataset for event-based video deraining, which includes rainy videos and temporally synchronized event streams. Moreover, large-scale synthetic datasets including various rains are also built.
\item We achieve superior performance over existing
state-of-the-art methods on both synthetic and self-collected real-world datasets. 
\end{itemize}

\section{Related Work}


\subsection{Single Image Deraining}
The single image deraining methods can be roughly divided into model-based methods and deep learning-based methods. Most of the model-based methods are proposed to utilize the intrinsic properties of the rain signal and the background texture details for separating the rain and background layers, e.g., discriminative sparse coding \cite{luo2015removing}, dictionary learning \cite{kang2011automatic}, non-local mean filter \cite{kim2013single}, Gaussian mixture model \cite{li2016rain}. Compared with the model-based methods, deep learning-based methods achieve better performance. Fu \textit{et al.} \cite{fu2017clearing, fu2017removing} proposed a guide filter to remove rain streaks from the high-frequency parts of the rainy image and directly predict the residual rain layer. Later works \cite{ahn2021eagnet, jiang2020decomposition} focused on designing more effective and advanced network architectures to obtain better performance.

\subsection{Video Deraining}
Video deraining is a long-standing ill-posed and challenging problem.
In contrast to single-image deraining \cite{ahn2021eagnet, zamir2021multi,jiang2023multi,zheng2022single}, temporal correlation and motion contexts can be additionally incorporated for video deraining. 
In \cite{garg2007vision}, Garg and Nayer first introduced the video deraining problem and detailed the properties of rain, such as the physical properties, spatial distribution, and appearance model.
Based on the unique properties of rain, model-based methods have been proposed to approach the video deraining task by utilizing more intrinsic priors to identify and remove rain in video. 
For example, chromatic properties \cite{liu2009pixel, zhang2006rain}, shape characteristics \cite{bossu2011rain, brewer2008using}, high frequency structure \cite{barnum2010analysis} are comprehensively explored for removing rains.
However, for heavy rain and other complex outdoor scenes, the above prior knowledge is not enough to support these model-based methods to identify the rain and background.

After the advent of deep learning, video deraining performance has been significantly improved. In \cite{chen2018robust}, Chen \textit{et al.} employed super-pixel segmentation to decompose the scene and then aligned the scene content at the super-pixel level.
Then a CNN was utilized to compensate for the misalignment and missing details. 
Yang \textit{et al.} \cite{yang2019frame} built a two-stage recurrent network that employs dual-level flow to regularize the learning process and predicted the rain-related variables in the video. 
In \cite{yang2021recurrent}, a new rain model considering rain accumulation, rain streaks, and rain occlusion was proposed.
Besides, a convolutional LSTM network was designed to make full use of the spatio-temporal redundancy. 
More recently, Yue \textit{et al.} \cite{yue2021semi} proposed a semi-supervised video deraining method.
They employed a dynamic rain generator to fit rain layers and took the real rainy videos into consideration for better performance in the real cases, substantially promoting deraining performance. 
In \cite{xue2021gta}, Xue \textit{et al.} designed a multi-stream coarse temporal aggregation module and a single-stream fine temporal aggregation module, 
which replaces the time-consuming alignment module to utilize the abundant temporal information.
Although these methods achieve considerable performance, they always pay great attention to complex architectures. In contrast, we resort to using an event camera which can provide helpful information for video deraining.

\subsection{Multi-Sensor Deraining}
Instead of using only one imaging sensor, some works \cite{zhang2020beyond, kim2014stereo} attempted to approach the deraining problem via building a stereo system, based on the observation that the effects of identical rain streaks across stereo images are different.
In \cite{kim2014stereo}, Kim \textit{et al.} warped the spatially adjacent right-view frame and subtracted the warped frame from the original frame.
Then a median filter was applied to the residual image for detecting rain streaks. 
Zhang \textit{et al.} \cite{zhang2020beyond} proposed the first semantic-aware stereo deraining network, which leverages semantic information and visual deviation between two views to detect and remove rain. 
Although stereo deraining methods promote deraining performance by taking the advantage of spatial-related information from stereo images, there still exist two common limits. 
First, in the scene with large and dense rain streaks, stereo images suffer from severe rain pollution and fail to provide adequate cues to each other.
Second, it is hard to model the temporal dynamics of rain for stereo sensors, which is more important for understanding the generation process and intrinsic property of rain layers.

\subsection{Event-Based Vision Techniques}

Event cameras have found extensive applications in various domains. Our work is closely related to prior research in event-based video deblurring \cite{jiang2020learning, wang2020event}, video super-resolution \cite{jing2021turning, han2021evintsr}, and video interpolation \cite{tulyakov2021time, yu2021training}. Notably, event cameras have recently been employed in video deraining \cite{suntip2023}, where a pioneering approach was introduced, leveraging multi-patch progressive learning for event-aware video deraining. In contrast, our method capitalizes on event data to explicitly detect rain and separate the rain layer, achieving better performance.

\begin{figure*}[t]
  \vspace{5.5mm}
  \centering
  \setlength{\abovecaptionskip}{2mm}
  \includegraphics[width=\linewidth]{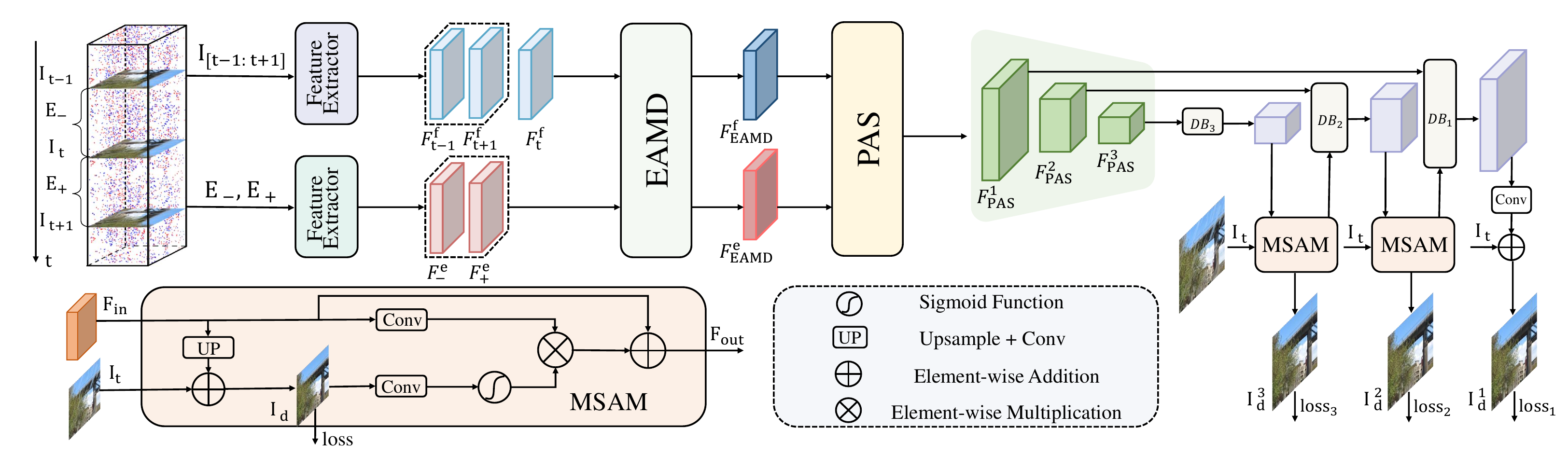}
  \caption{Pipeline of our event-guided video deraining method.
  \textbf{$I_t$, $I_{t-1}$, $I_{t+1}$, $E_{-}^{t}$, $E_{+}^{t}$} denote the target frame, its neighboring frames and in-between events.
  $EAMD$ and $PAS$ denote the event-aware motion detection module and pyramidal adaptive selection module respectively.
  The pyramidal features in the light green trapezoid region are the outputs of $PAS$.
  The reconstruction module is composed of three decoder blocks ($DB_{i}$) and two $MSAMs$. The output of the decoder block ($DB_{i}$) is the decoded rain feature and $I_{d}^i$ denotes the derained frame. 
  We build $DB_{i}$ by using eight convolutional blocks and present the details of $MSAM$ in the lower left part.
  }
  \label{fig:overview}
  \vspace{3mm}
\end{figure*}

\section{Preliminaries}

\subsection{Rain in Frame Cameras}

There are many unique properties of rain, such as geometric property \cite{beard1987new}, chromatic property \cite{zhang2006rain}, spatial and temporal property \cite{zhang2006rain}.
In our work, we focus on the photometry of rain \cite{garg2007vision}.

In \cite{garg2007vision}, Garg and Nayer pointed out that a raindrop acts as a spherical lens that refracts and reflects light, and produces a positive intensity change at a pixel.
The imaging process is formulated as:
\begin{equation}
\label{framecamera_rain}
I_{r}(\vec{x})=\int_{0}^{\tau} E_{r} dt+\int_{\tau}^{T} E_{b} dt,
\end{equation}
where the $\tau$ is the time that a drop projects onto a pixel, $T$ is the exposure time of the camera.
We can see that the intensity for pixel $\vec{x}$ is a linear combination of raindrop irradiance $E_r$ and background irradiance $E_b$, resulting in a fused measurement that is hard to separate.
Due to the intrinsic property of capturing at a fixed internal for the frame camera, severely motion-blurred rain streaks are produced by high-speed moving raindrops.
Moreover, the motion priors out of exposure time are inaccessible, leading to inevitable performance degeneration.

\subsection{Rain in Event Cameras}

The event camera outputs a sparse data stream $\mathcal{E} = \{ e_k \}_{k=1}^{N_e}$, where $N_e$ is the number of events, reporting the intensity changes in the scene.
Each triggered event can be represented as a quaternion $(x_k, y_k, t_k, p_k)$, describing the spatial coordinates, timestamp and polarity respectively.
Each pixel of an event camera is able to independently and asynchronously produces an event if intensity change reaches a threshold:

\begin{equation}
\log I(\textbf{x}, t)-\log I(\textbf{x}, t-\Delta t)=\pm C,
\end{equation}
where $I(\textbf{x},t)$ is the intensity of pixel {\textbf{x}} at time t, C is the contrast threshold that can be obtained from the camera configuration, $\Delta t$ is the time interval of the last event triggered at the same position. 
The survey paper \cite{gallego2020event} provides more details of the event camera.
In a rain scene, the output of an event camera can be given as:
\begin{equation}
\label{eventcamera_rain}
    \mathcal{E} = \mathcal{E}_r\cup \mathcal{E}_b,
\end{equation}
where $\mathcal{E}_r$ and $\mathcal{E}_b$ are event streams triggered by motions of rain and background respectively.

Particularly, there are two differences between a frame camera and an event camera for rain imaging.
(\textbf{i}) The measurement of an event camera (Equation (\ref{eventcamera_rain})) only focuses on the motion regions, while a frame camera (Equation (\ref{framecamera_rain})) records motion and static regions simultaneously;
(\textbf{ii}) An event camera outputs data at microseconds that approximates temporally-continuous recording, while a frame camera only produces data within the exposure time.
The above observations motivate us to approach video deraining with an event camera, by exploiting its perception of motion variation and microsecond temporal resolution property.

\begin{figure*}[t!]
  \vspace{5.5mm}
  \centering
    \includegraphics[width=\linewidth]{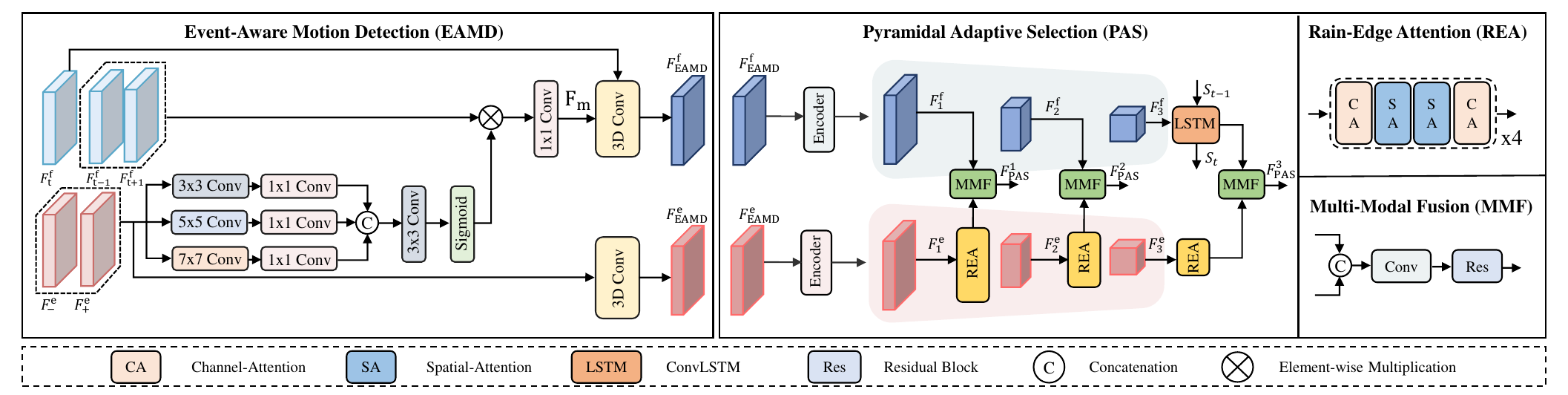}
  \caption{Detailed architecture of event-aware motion detection ($EAMD$) module and pyramidal adaptive selection ($PAS$) module. $EAMD$ first selectively extracts the motion-aware information from the neighboring frame features with the guidance of the motion-attention map which is learned from the event features, then employs a 3D convolution block to fuse the detected motion features from the neighboring frames with the central target frame features. $PAS$ adopts the pyramidal architecture to embed the multi-scale features in the frame domain and event domain respectively. At each scale, the rain-edge attention ($REA$) block is designed to enhance the motion of rain layers in the event domain and the multi-modal fusion ($MMF$) block is employed for effectively fusing multi-modal features from frames and events. Additionally, a ConvLSTM layer is employed at the last scale encoder to model the long-term correlation of frames.}
  \label{fig:EAMD}
\end{figure*}

\section{Proposed Method}


We propose a learning-based network, named as \textbf{EGVD}, for approaching video deraining with an event camera.
As shown in Fig. \ref{fig:overview}, our EGVD comprises four components, i.e., a feature extractor, an event-aware motion detection (EAMD) module, a pyramidal adaptive selection (PAS) module, and a reconstruction module, which tightly collaborate for video deraining.

\subsection{Overview}
We consider three consecutive frames and in-between events as the input.
In specific, given a target frame $I_t$ and its neighboring frames $I_{t-1}, I_{t+1}$, the events between $I_{t-1}$ and $I_{t}$ and the events between $I_{t}$ and $I_{t+1}$ are converted into event voxel grids $E_{-}$ and $E_{+}$ respectively. The conversion method will be described in Section \ref{event_represent}.

We then utilize the feature extractors to extract the features from three frames $I_{[t-1:t+1]}$ and two event voxel grids $E_{-}$, $E_{+}$, forming frame features $F_{[t-1:t+1]}^{f} \in \mathbb{R}^{C\times 3\times H\times W}$ and event features $F_{-}^{e} \in \mathbb{R}^{C\times 1\times H\times W}$, $F_{+}^{e} \in \mathbb{R}^{C\times 1\times H\times W}$ respectively.

The feature extractors of frame and event share a similar network architecture, which contains one convolution layer and one residual block. 
After feature extraction, we feed the frame features $F_{[t-1:t+1]}^{f}$ and the event features $F_{-}^{e}$, $F_{+}^{e}$ to EAMD, which is responsible for detecting and enhancing the motion-aware information of frame features with the guidance of events.
Then, the enhanced frame feature $F_{EAMD}^{f}$ and event feature $F_{EAMD}^{e}$ from EAMD are fed into PAS for motion separation and multi-modal fusion.
Finally, the output features of PAS pass through a reconstruction module for predicting the residual layer, to which the target degraded frame is added for obtaining the derained frame.

\subsection{Event Representation}
\label{event_represent}

Compared with conventional frame data, event data is essentially a kind of sparse spatio-temporal stream.
We first convert it into a fixed-size representation.
Specifically, 
we opt to encode event data that is triggered in the time interval between two adjacent frames in a spatio-temporal voxel grid, sharing a similar idea with \cite{zhu2019unsupervised}.
Given an event stream $\mathcal{E}=\{ e_k = (x_k, y_k, t_k, p_k) \}_{k=0}^{N_e-1}$ with a duration $\Delta T=t_{N_e-1}-t_{0}$, we uniformly divide the duration $\Delta T$ into $B$ time bins.
In this way, every event distributes its polarity to two temporally closet voxels. Mathematically, the event voxel grid is formulated as:
\begin{equation}
\label{equa:event_repre}
E\left(x_{m}, y_{n}, t_{l}\right)=\sum_{\substack{(x_k,y_k) = (x_m, y_n) \\ k\in\{0,\cdots, N_e-1\}}} p_{k} \max \left(0,1-\left|t_{l}-t_{k}^{*}\right|\right),
\end{equation}
where $t_{k}^{*} \triangleq \frac{B-1}{\Delta T}\left(t_{k}-t_{0}\right)$ is the normalized event timestamp, and $t_l \in \{0,\cdots,B-1\}$ denotes the index of time bin. 


\subsection{Event-Aware Motion Detection}
Unlike single-image deraining, additional temporal information that exists across adjacent frames can be exploited for video deraining. 
However, directly packaging the multi-frame features into a network is not necessarily effective, but burdens the rain removal task due to the introduction of excessive redundant information. 
In contrast, we selectively extract the motion-aware information of neighboring frames. 
The motion information typically acts as favorable clues for modeling the dynamic generation process of rain layers, which is able to identify the rain region. 
However, it is difficult to obtain motion-aware information for conventional frame cameras, which are limited by their low temporal resolution and low dynamic range properties of imaging.
Fortunately, event cameras are capable of accurately detecting motion variations even for large motion and low light scenes thanks to their unique properties.
Therefore, we design an event-aware motion detection module to detect the motion features from the neighboring frames and then employ a 3D convolution block to fuse them with central target frame features.

\begin{figure}[t]
  \centering
  \includegraphics[width=\linewidth]{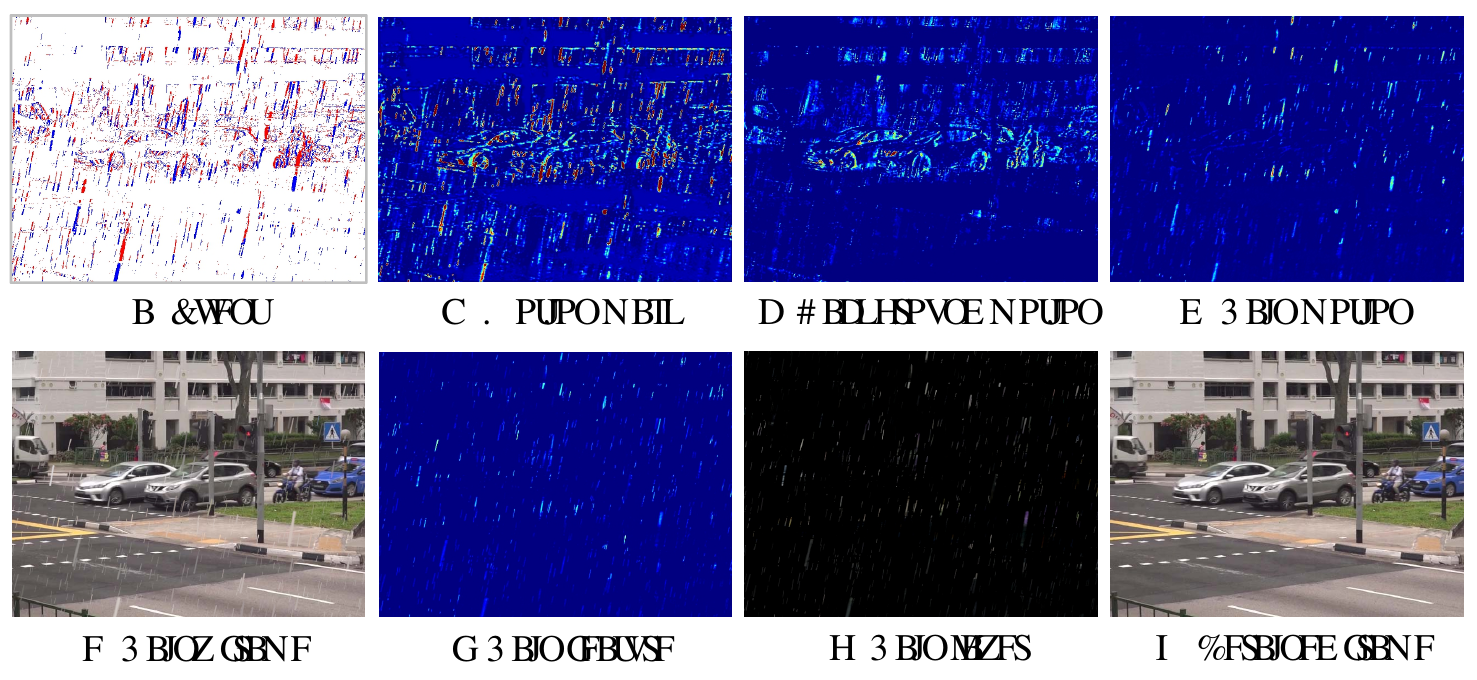}
  \caption{
  Visualizations of (a) the last channel map of the event voxel grid (blue: positive event; red: negative event), (b) motion mask $M$ in $EAMD$ module, (c, d) background and rain features outputted by $REA$ of $PAS$ module (two from all channel maps), (e) rainy frame, (f) rain features outputted by the last decoder block ($DB_{1}$ in Fig.~\ref{fig:overview}), (g) estimated rain layer, and (h) derained frame. Zoom-in for better visualization.}
  \label{fig:motion}
\end{figure}

More specifically, as shown in Fig. \ref{fig:EAMD}, we first generate a motion-attention mask $M$ using event features $F_{-}^{e}$, $F_{+}^{e}$.
We formulate it as:
\begin{equation}
\begin{aligned}
M =& \sigma\left(F_{m}^e\right), \\
F_{m}^e =& \psi _ { 3 \times 3 } \left(\left[\left(\psi _ { 1 \times 1}\left(\psi_{7 \times 7}\left(F^{e}\right)\right)\right), \right.\right.\\
&\left.\left.\left(\psi_{1 \times 1}\left(\psi_{5 \times 5}\left(F^{e}\right)\right)\right), \left(\psi_{1 \times 1}\left(\psi_{3 \times 3}\left(F^{e}\right)\right)\right)\right]\right), \\
F^e =& [F_{-}^{e}, F_{+}^{e}],
\end{aligned}
\end{equation}
where $M$ is the predicted motion-attention map, $\sigma$ is the sigmoid activation function which restricts the outputs in (0,1), and $[\cdot]$ indicates channel-wise concatenation. 
We adopt seven convolution layers with different kernel-sizes, i.e., $1\times 1$, $3\times 3$, $5\times 5$, $7\times 7$, denoted as $\psi_{1 \times 1}, \psi_{3 \times 3}, \psi_{5 \times 5}, \psi_{7 \times 7}$, to extract the information of different receptive fields from event features, which enables the effective detection of motion-aware regions.
We visualize the motion-aware mask $M$ in Fig. \ref{fig:motion} (b).
As can be seen, the motion regions of rain and moving objects can be clearly detected.

Afterwards, we rectify the frame features $F_{t-1}^{f}$, $F_{t+1}^{f}$ using the predicted motion-aware mask $M$:
\begin{equation}
\begin{aligned}
F_m = \psi _ { 1 \times 1 }\left(M \otimes \left[F_{t-1}^f, F_{t+1}^f\right]\right), 
\end{aligned}
\end{equation}
where $\otimes$ stands for the element-wise multiplication. 
Then a 3D convolution layer is adopted to aggregate the multi-frame contexts and further enhance the motion-aware information in the frame domain, generating the enhanced frame features $F^f_{EAMD}$.
Similarly, in the event domain, we also aggregate the event features using a 3D convolution layer, yielding the enhanced event features $F^e_{EAMD}$.
These two 3D convolution layers do not share parameters.
We formulate them as:
\begin{equation}
\begin{aligned}
F^f_{EAMD}, F^e_{EAMD} = \psi_{3D}\left([F^{f}_{t}, F_m]\right), \psi_{3D}\left( F^{e}\right).
\end{aligned}
\end{equation}

\subsection{Pyramidal Adaptive Selection}
One key challenge of video deraining is how to accurately separate the rain layer and background layer, which are typically blended in the image domain and feature domain.
It is difficult to achieve with frame cameras, especially for heavy rain scenes.
Thanks to the properties of high temporal resolution and high dynamic range, event cameras are able to enhance the visibility of rain and moving objects of background in the event domain, hence providing strong guidance for rain-background separation.
To this end, we build a pyramidal adaptive selection module as illustrated in Fig. \ref{fig:EAMD}.

To be specific, we first adopt the encoder architecture of standard UNet \cite{ronneberger2015u} to embed deep multi-scale features in the frame domain and event domain respectively.
We denote the extracted features as $F_i^f$ and $F_i^e$, where $i \in \{1, 2, 3\}$ denotes the scale index.
At each scale, we design a rain-edge attention (REA) block to enhance the motion of rain layers in the event domain and employ a multi-modal fusion (MMF) block to adaptively fuse the information of both frame and event modalities. 
We implement our REA block by adopting four symmetric channel-spatial-spatial-channel attentions and concatenate the multi-modal features before feeding to a residual convolution to form an MMF block. 
Moreover, in order to model the long-term correlation of frames, we also employ a ConvLSTM layer at the last scale encoder.

With the design of repeated symmetric channel-spatial-spatial-channel attentions, the REA block is able to cyclically learn the spatial/channel importance in the event domain.
In this way, the rain and background motions are able to be separated effectively, which can be clearly observed in Fig. \ref{fig:motion} (c), (d).
The fused multi-modal features of the MMF block take the complementary merits of multi-modal features from frames and events, favorably promoting the video deraining performance.
The computing process of the PAS module can be formulated as:
\begin{equation}
\begin{aligned}
    F_{PAS}^i = \left\{ 
    \begin{aligned}
    MMF\left(REA\left(F_i^e\right), F_i^f\right), i=1,2\\
    MMF\left(REA\left(F_i^e\right), LSTM\left(F_i^f, S_{t-1}\right)\right), i=3
    \end{aligned}
    \right.
\end{aligned}
\end{equation}
where $i \in \{ 1, 2, 3 \}$ denotes the scale index, and $S_{t-1}$ is the previous state of the ConvLSTM layer.

\subsection{Rain Layer Reconstruction}

Instead of predicting a clean background layer, our network predicts a negative rain layer, which can be attributed to two reasons. 
On one hand, the rain layer is sparser than the background layer, making it easier for the network to converge, which has been proved in \cite{fu2017removing}. 
On the other hand, in our event-guided video deraining setting, we separate the rain layer and background layer from the perspective of motions. 
The moving edges of rain help us model the dynamics and spatial distribution of rain streaks, which makes it easier to directly predict the rain layer. 
Typically, most deraining methods choose to predict the rain layer for the first reason, however, due to the specificity of our setting, the second reason is more important. 
We also validate the effectiveness of the way to directly predict the rain layer in Section \ref{ex:ablation}.

As shown in Fig. \ref{fig:overview}, given the features $\{ F_{PAS}^i \mid i \in \{ 1, 2, 3\} \}$ generated by PAS, we use three convolutional blocks to progressively reconstruct the negative rain layer.
Inspired by \cite{zamir2021multi}, we also build a multi-scale supervised attention module (MSAM) to enhance the feature learning with the supervision of a ground-truth clean frame.
After obtaining the negative rain layer, we add it with the input target rainy frame to obtain the clean background.
The rain layers at three scales will be supervised during the training phase, and we select the output of the last scale as the final reconstructed rain layer when inference.

\subsection{Loss Function}
\label{sec:loss}
We apply the negative SSIM loss for computing the distance between the intermediate prediction at each scale and the ground-truth frame.
The overall loss is the sum of losses at different scales, which is formulated as:

\begin{equation}
\mathcal{L}=- \sum_{i=1}^{3} S S I M\left(I_{d}^i, I_{gt}\right),
\end{equation}
where $I_{d}^i$, $I_{gt}$, $i$ indicates the derained frame, corresponding ground-truth frame and scale index, respectively.

\section{Experiments}

\subsection{Experimental Settings}
\subsubsection{Synthetic Datasets}
To the best of our knowledge, there are no benchmark datasets that provide rainy videos with temporally synchronized event streams and corresponding ground-truth clean videos. 
Hence, we generate large-scale synthetic datasets for event-based video deraining.
Specifically, we first use a video editing software to synthesize rain streaks. 
We randomly set the parameters, e.g., scale, density, wind direction, camera shutter speed, scene depth and capacity.
Afterwards, we choose some video clips as ground-truth, which are overlaid by the generated rain layers to produce rainy videos.
We finally choose the open event simulator \cite{rebecq2018esim} to generate event streams from rainy videos.
In such a way, we generate four synthetic datasets in total. 
We name them following the pattern ``N-D'', where ``N'' indicates Neuromorphic and ``D'' will be replaced with the name of the original dataset from which the clean videos come. The details of four synthetic datasets are provided below.

\textit{N-NTURain}: It is generated from 16 rain-free sequences in the NTURain \cite{chen2018robust} dataset, which is widely used as a benchmark for video deraining methods, but with re-synthesized rain streaks. 
Before synthesizing rain streaks, we use the interpolation method proposed in \cite{jiang2018super} to increase the frame rate for better event simulation. 
We adopt the default dataset splitting as in \cite{chen2018robust}.
For training, we synthesize 3 to 4 different rain appearances over each clean video, resulting in 25 training videos.
For testing, we produce 8 videos with varying rain parameters.

\textit{N-GoproRain}: 
To take the more complex motion information into consideration and further validate the ability of motion separation of $REA$,
we choose GoPro \cite{Nah_2017_CVPR} that is built by a high-speed camera for dynamic scene deblurring as the clean video source.
GoPro is a more challenging dataset that contains a variety of object motions and camera motions.
We adopt the default dataset splitting as in \cite{Nah_2017_CVPR}.
For training, we synthesize 3 different rain appearances over each clean video, resulting in 66 training videos.
For testing, we produce 11 videos with varying rain parameters.

\textit{N-AdobeRainH, N-AdobeRainL}: Similar to RainSynLight25/RainSynComplex25 \cite {liu2018erase} and Rain100L/H \cite{yang2017deep}, we synthesize two datasets containing only heavy and light rain layers based on Adobe240fps \cite{su2017deep}, which are captured outdoors at 240fps. 
Each of them contains 109 training video clips and 19 testing video clips. 


\subsubsection{Real-World Dataset}
For real-world evaluation, we construct a real-world dataset using a Color-DAVIS346 camera \cite{taverni2018front}. 
This camera has a high-speed event sensor and a low frame-rate active pixel sensor (APS) with a resolution of $260 \times 346$, which produces event streams and low frame-rate frames.
We capture rainy videos in real rainy days.
By doing so, we are able to model the real spatio-temporal distribution of rain streaks along with the light conditions of real scenes.
Moreover, we collect the data at different times and carefully control exposure time to obtain the rainy videos with different types of rain streaks under different lighting conditions. 
Our real-world dataset consists of two groups of data sets, the videos in the first group are captured by a still camera, while those in the second group are captured by a panning camera. Each of them varies from light drizzling to heavy falls. In total there are 10 video clips in our real-world dataset.
We name our real-world dataset Rain-DAVIS. 

\begin{table*}[t!]
\centering
\setlength{\tabcolsep}{10pt}
\renewcommand{\arraystretch}{1.2}
\caption{Quantitative comparisons on four synthetic datasets. Best in bold, the runner-up with an underline.
}
 \label{tab:benchmarking}
\resizebox{0.95\linewidth}{!}{
\begin{tabular}{ccccccccccccc}
\toprule[0.8pt]
\multirow{2}{*} {Methods} & \multicolumn{2}{c}{N-NTURain}    && \multicolumn{2}{c}{N-GoProRain}  && \multicolumn{2}{c}{N-AdobeRainL} && \multicolumn{2}{c}{N-AdobeRainH} \\
\cline{2-3} \cline{5-6} \cline{8-9} \cline{11-12}
 & PSNR$\uparrow$ & SSIM$\uparrow$ && PSNR$\uparrow$ & SSIM$\uparrow$ && PSNR$\uparrow$ & SSIM$\uparrow$ && PSNR$\uparrow$ & SSIM$\uparrow$ \\ 
\hline
DualResNet \cite{liu2019dual}          & 37.36             & 0.9763              && 28.30             & 0.8672             && 31.59            & 0.9356             && 27.81               & 0.8804          \\
PReNet \cite{ren2019progressive}               & 37.64             & 0.9803              && 33.72             & 0.9652             && 38.55             & 0.9889             && 32.29              & 0.9551          \\
DCSFN \cite{wang2020dcsfn}               & 39.30             & 0.9865  && 36.96             & 0.9808             && 42.18  & \underline{0.9944} && 34.46              & 0.9695          \\
MPRNet \cite{zamir2021multi}               & 39.83 & \underline{0.9875}             && 38.07  &  \underline{0.9851}    && 41.24             & 0.9939              && 34.74  & 0.9746    \\
SAVDTDC \cite{yan2021self}                 & 34.96             & 0.9708             && 30.41             & 0.9360             && 36.09            & 0.9832               && 29.03              & 0.9194          \\
S2VD \cite{yue2021semi}                 & 36.30             & 0.9718             && 25.43             & 0.8346             && 33.38            & 0.9505               && 26.89              & 0.8548          \\
RMFD \cite{yang2021recurrent}                 & 36.04             & 0.9746             && 30.40             & 0.9333             && 36.23            & 0.9819               && 30.08              & 0.9283          \\
GTA-Net \cite{xue2021gta}               & 38.88             & 0.9852             && 35.84             & 0.9761             && 40.43             & 0.9917               && 33.74              & 0.9632          \\ 
DANet \cite{ijcai2022p137}               & 36.71             & 0.9798             && 33.31             & 0.9642             && 37.17             & 0.9871               && 31.75             & 0.9515          \\ 
DFTL-X \cite{ijcai2022p205}               & 39.40             & 0.9860             && 37.91             & 0.9831             && 41.48             & 0.9938               && 34.53              & 0.9701          \\ 
Restormer \cite{zamir2022restormer}               & 39.49             & 0.9861             && \underline{38.26}             & 0.9843             && 40.12             & 0.9918               && \underline{36.43}              & \underline{0.9790}          \\
EAVD \cite{suntip2023}               & \underline{39.94}             & 0.9832             && 34.30             & 0.9609             && \underline{42.59}            & 0.9931               && 34.69              & 0.9614          \\ 
EGVD (Ours)           & \textbf{42.20} & \textbf{0.9903}      && \textbf{39.59}   & \textbf{0.9874}    && \textbf{45.67} & \textbf{0.9964} && \textbf{38.20} & \textbf{0.9815} \\ 
\bottomrule[0.8pt]
\end{tabular}
}
\end{table*}
\begin{figure*}[t!]
  \centering
  \includegraphics[width=0.9\linewidth]{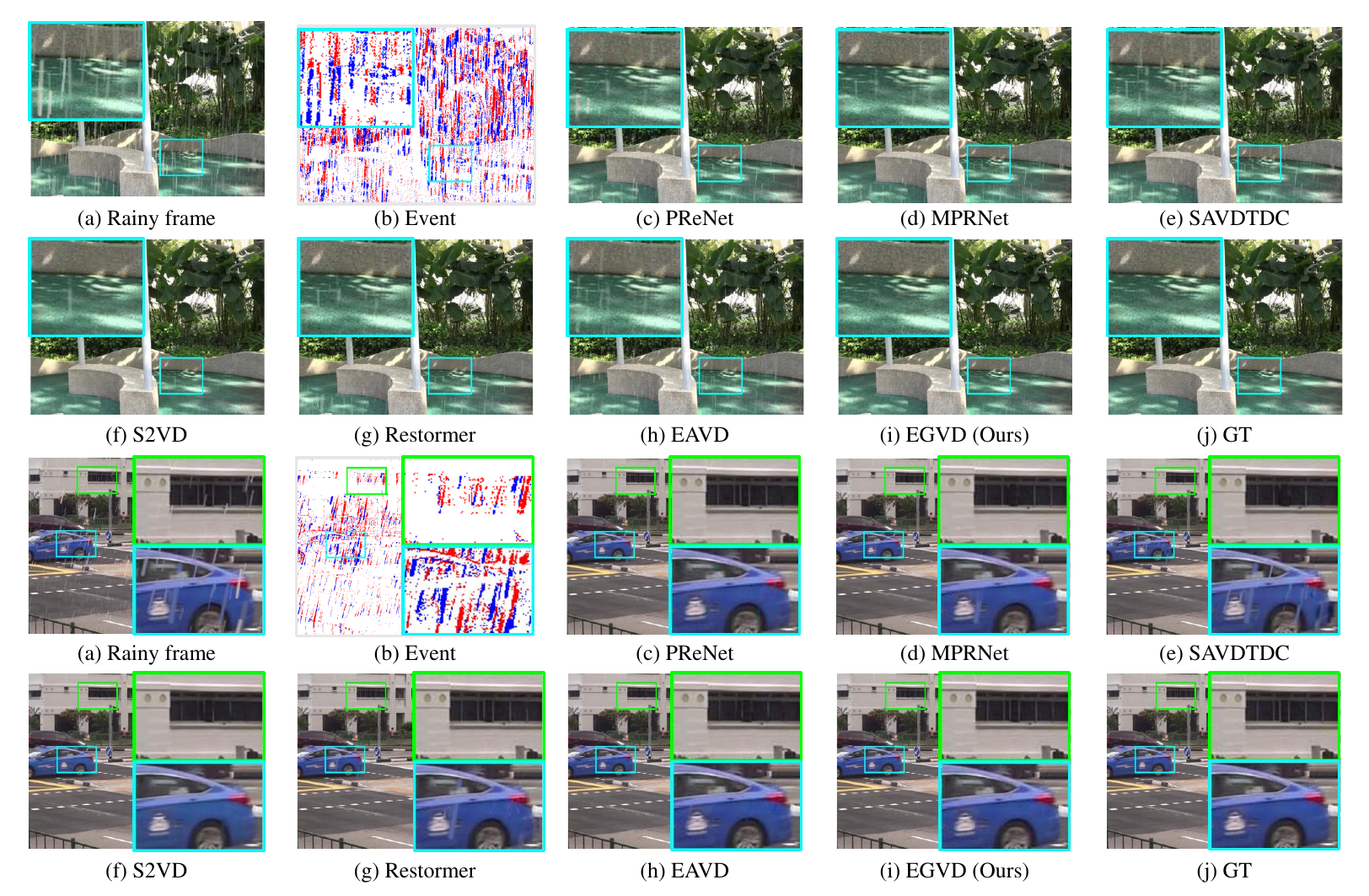}
  \caption{Visual comparisons on N-NTURain. We visualize the last channel map of the event voxel grid (blue: positive event; red: negative event) in (b). Zoom-in for better visualization.}
  \label{fig:qualitative_syn_v1}
\end{figure*}

\begin{figure*}[t!]
  \centering
  \includegraphics[width=0.9\linewidth]{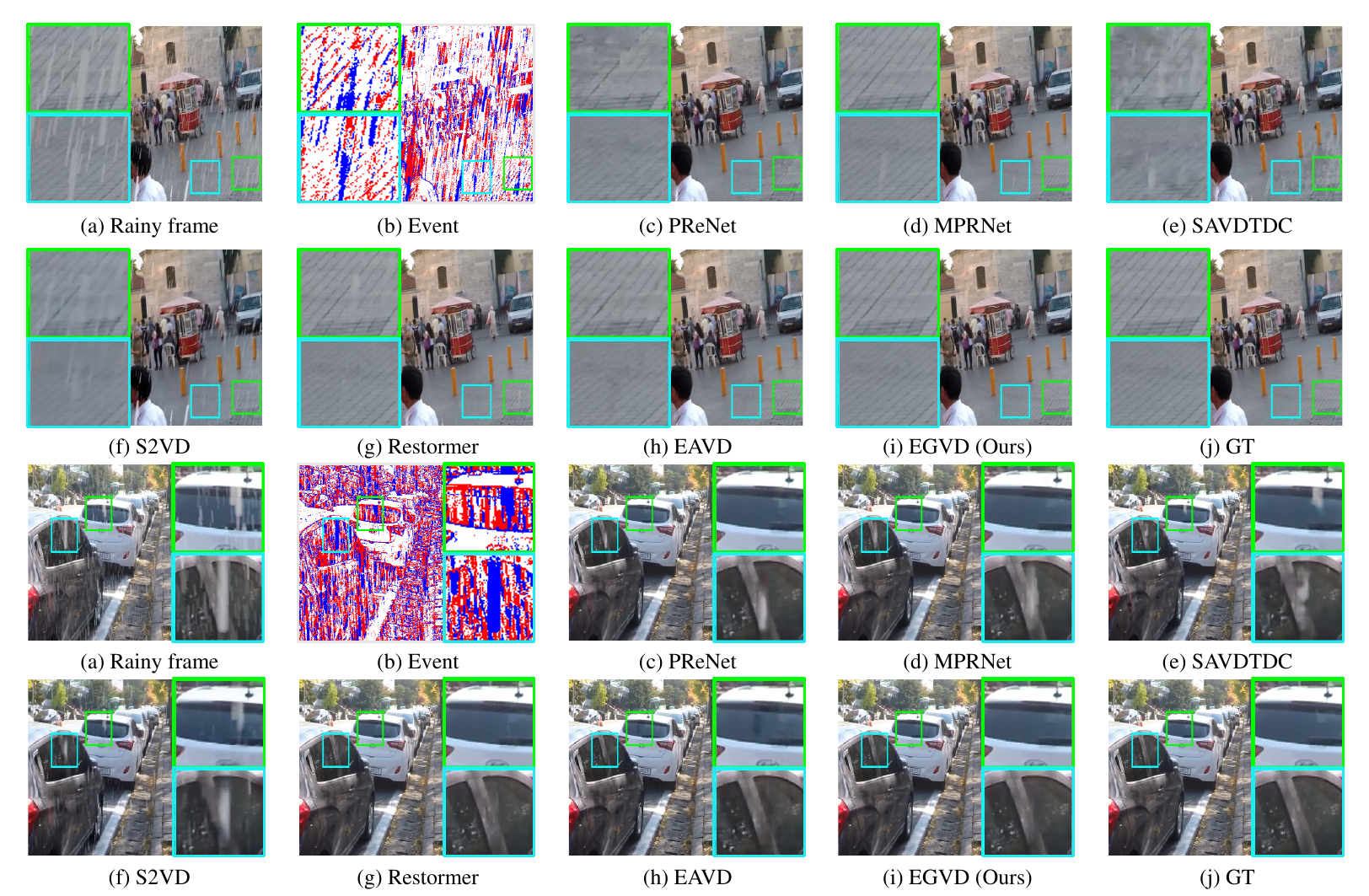}
  \caption{Visual comparisons on N-GoproRain. We visualize the last channel map of the event voxel grid (blue: positive event; red: negative event) in (b). Zoom-in for better visualization.}
  \label{fig:qualitative_syn_v2}
\end{figure*}

\begin{figure*}[t!]
  \centering
  \includegraphics[width=0.9\linewidth]{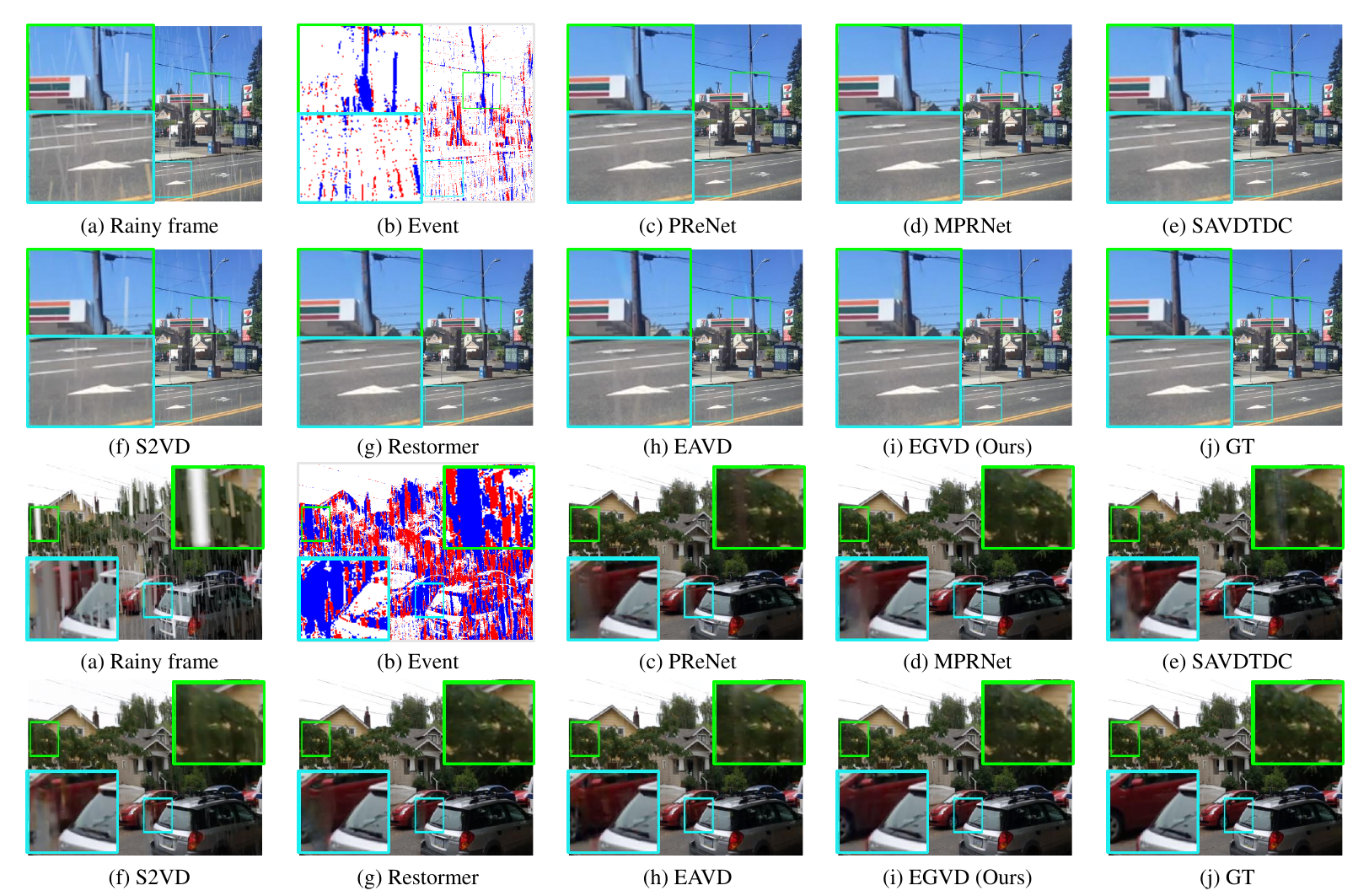}
  \caption{Visual comparisons on N-AdobeRainL and N-AdobeRainH. We visualize the last channel map of the event voxel grid (blue: positive event; red: negative event) in (b). Zoom-in for better visualization.}
  \label{fig:qualitative_syn_ll}
\end{figure*}

\subsubsection{Implementation Details}
We train our network on random-cropped 128x128 patches with a batch size of 2 for 500 epochs. 
We use Adam optimizer \cite{kingma2014adam} with the initial learning rate of $1 \times 10^{-4}$, which is steadily decreased to $1 \times 10^{-5}$ using the cosine annealing strategy \cite{loshchilov2016sgdr}. 
We implement our deep model using PyTorch 1.1 \cite{paszke2019pytorch} and conduct all experiments on an NVIDIA GTX1080Ti GPU.

\subsection{Comparisons with State-of-The-Art Methods}

\subsubsection{Baselines}
We make comparisons with the state-of-the-art video deraining methods: SAVDTDC \cite{yan2021self}, S2VD \cite{yue2021semi}, RMFD \cite{yang2021recurrent}, GTA-Net \cite{xue2021gta} and single image deraining methods: MPRNet \cite{zamir2021multi}, DCSFN \cite{wang2020dcsfn}, PReNet \cite{ren2019progressive}, DualResNet \cite{liu2019dual}. The event-based video deraining method EAVD \cite{suntip2023} is also evaluated for comparison.
Because our method is trained in a supervised manner, so we follow the instructions in \cite{yue2021semi} for training a supervised S2VD \cite{yue2021semi} using ground-truth data for a fair comparison. 
For training other baselines, we carefully follow the training strategy provided by the authors. The average PSNR and SSIM are used as evaluation metrics. 

\begin{figure*}[t!]
  \centering
  \includegraphics[width=0.9\linewidth]{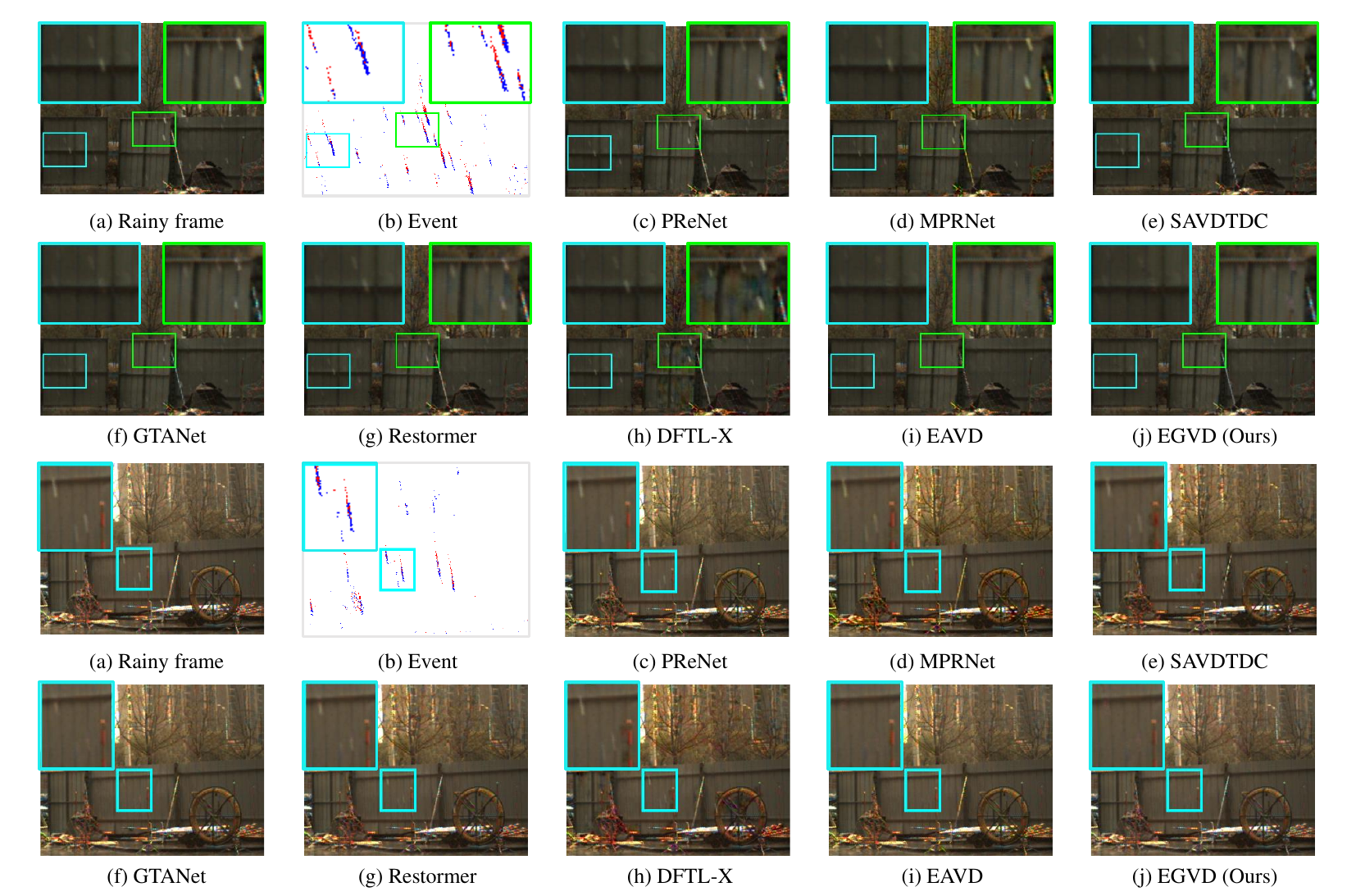}
  \caption{
  Visual comparisons on our real-world dataset Rain-DAVIS. We visualize the last channel map of the event voxel grid (blue: positive event; red: negative event) in (b).
  Our method is able to remove the rain streaks and restore clearer texture information, while other methods cannot remove some rain streaks and lose details of non-rain regions. Zoom-in for better visualization.}
  \label{fig: qualitative results on real}
  \vspace{2mm}
\end{figure*}

\subsubsection{Results on Synthetic Datasets}
Table~\ref{tab:benchmarking} lists the average PSNR and SSIM results on four synthetic datasets, including N-NTURain, N-GoproRain, N-AdobeRainL, and N-AdobeRainH.
Evidently, EGVD attains the best performance. 
Especially on N-NTURain, N-AdobeRainL, and N-AdobeRainH, EGVD achieves at least 2 dB PSNR gains, which demonstrates that our method obtains better performance in removing rain streaks and preserving clean texture details.
This could be attributed to its powerful capability in effectively utilizing the motion features from the neighboring frames and accurately modeling the spatio-temporal distribution of rain layers with the guidance of event data.
The probabilistic video deraining method S2VD \cite{yue2021semi} is less effective in these datasets because it adopts a dynamical rain generator consisting of a transition model and an emission model to represent the dynamics of rains and mimic the generation process of rain layers based on the statistics of rain streaks.
It fails to handle complex and heavy rain in these synthetic datasets. With the additional event information, EAVD performs well in N-NTURain and N-AdobeRainL, but relatively worse in the other two datasets. The possible reason is that the scenes in N-GoProRain and N-AdobeRainH usually have heavy rain, which might be hard for EAVD to tackle with.

Visual comparison results on the N-NTURain dataset are depicted in Fig. \ref{fig:qualitative_syn_v1}. In the first example, state-of-the-art video deraining techniques such as S2VD \cite{yue2021semi} and SAVDTDC \cite{yan2021self} outperform single-image deraining methods like MPRNet \cite{zamir2021multi}, Restormer \cite{zamir2022restormer}, and PReNet \cite{ren2019progressive} due to their incorporation of additional temporal consistency and correlation. However, they exhibit shortcomings in detecting and removing certain small rain streaks. EAVD \cite{suntip2023} also demonstrates commendable deraining results, albeit with residual rain traces. In contrast, EGVD effectively harnesses contextual information from adjacent frames, accurately identifying rain streaks from a motion perspective. Consequently, EGVD excels in rain streak removal and approaches ground truth quality. In the second example, the frame-based deraining methods struggle to distinguish rain streaks from the background layer, leading to either missed detections or excessive smoothing of non-rain regions resembling rain streaks. Conversely, the event camera, capturing motion information, guides our method in correctly identifying rain streaks and non-rain regions. Consequently, our approach efficiently eliminates rain streaks while preserving clean texture details. To further bolster our claims, we present additional visual comparisons on the N-GoproRain, N-AdobeRainH, and N-AdobeRainL datasets, displayed in Fig. \ref{fig:qualitative_syn_v2}, \ref{fig:qualitative_syn_ll}, respectively.


Clearly, our EGVD method outperforms other deraining techniques across various rain intensities, from light drizzles to heavy downpours. It excels at effectively eliminating rain streaks and accurately restoring intricate texture details, whereas the compared deraining methods struggle to completely eradicate the rain streaks and often incorrectly handle background layer details.

\subsubsection{Generalization on Real-World Dataset}
We evaluate the generalization capability of representative methods using our real-world dataset, Rain-DAVIS. To ensure a fair comparison, we employ pre-trained models on the N-NTURain dataset to eliminate real rain streaks in Rain-DAVIS. Visual results are presented in Fig. \ref{fig: qualitative results on real}. Frame-based methods struggle to detect and remove rain streaks due to differences in rain patterns between synthetic and real rainy videos. Consequently, frame-based rain removal methods yield deraining results with incomplete rain streak removal and some detail loss. In contrast, our method, guided by events, excels in rain streak detection and removal, surpassing the performance of EAVD. As evident in the two examples, the compared methods produce deraining results with incomplete rain streak removal, while our approach successfully restores clearer rain-free frames and preserves fine details.
\begin{figure}[t!]
  \centering
  \includegraphics[width=\linewidth]{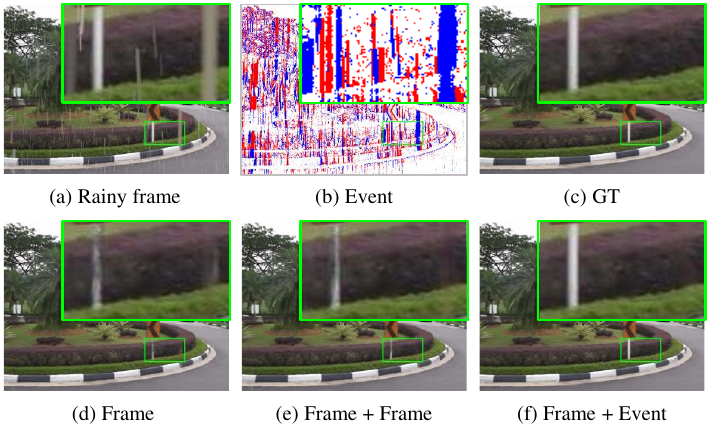}
  \caption{Visual results of ablation on input data. We visualize the last channel map of the event voxel grid (blue: positive event; red: negative event) in (b). Zoom-in for better visualization.}
  \label{fig: ablation on input data}
\end{figure}

\begin{table}[t!]
\centering
\setlength{\tabcolsep}{10pt}
\renewcommand{\arraystretch}{1}
  \caption{Ablation results on input data. We take the only frames as input (Frame), keep the event branch but replace its input with the frames (Frame+Frame), and take frames and events as input (Frame+Event) to investigate the influence of input data. The best results are marked in bold.}
  \label{tab:input}
  \resizebox{0.95\linewidth}{!}{
  \begin{tabular}{cccc}
    \toprule[0.8pt]
    Metrics &    Frame         &    Frame + Frame   &  Frame + Event   \\
    \midrule
    PSNR    &   37.33          &  36.39           &  \textbf{42.20}         \\
    SSIM    &   0.9794         &  0.9698          &  \textbf{0.9903}        \\
  \bottomrule[0.8pt]
\end{tabular}
}
\end{table}

\begin{table}[t!]
\centering
\setlength{\tabcolsep}{10pt}
\renewcommand{\arraystretch}{1.2}
\caption{Ablation Results on different modules. We replace the EAMD module with an equal-parameter convolutional block (Model\#A), replace the REA module with an equal-parameter convolutional block (Model\#B), disable the state propagation of ConvLSTM (Model\#C), keep all components (Model\#D) to investigate the influence of different modules. The best results are marked in bold.}
\label{tab:abl_module}
\vspace{2mm}
\resizebox{1\linewidth}{!}{
\begin{tabular}{ccccccc}
\toprule[0.8pt]
Variants  &  EAMD & REA & ConvLSTM && PSNR & SSIM \\
\hline  
Model\#A   & & $\checkmark$& $\checkmark$&&  41.84 & 0.9899 \\
Model\#B    &$\checkmark$ & & $\checkmark$&&  41.69 & 0.9889 \\
Model\#C    &$\checkmark$& $\checkmark$ & &&  42.10 & 0.9901 \\
Model\#D    &$\checkmark$ & $\checkmark$ & $\checkmark$ && \textbf{42.20} & \textbf{0.9903} \\
\bottomrule[0.8pt]
    \end{tabular}
}

\end{table}

\subsection{Ablation Study}
\label{ex:ablation}
We examine the efficacy of key components within our approach by a comprehensive series of ablation experiments. Specifically, we scrutinize the impact of various factors, including input data, network modules, mapping ways, and loss functions. All of these ablation experiments are conducted using the N-NTURain dataset as our testbed.

\subsubsection{Influence of Input Data}
We argue that the utilization of event data enhances the video deraining process. To substantiate the efficacy of our approach, we conduct three  experiments:
\textbf{1}) Only frame as input.
We remove the event branch (containing operations related to event data) in Fig. \ref{fig:overview}, thus only frames are taken as input.
\textbf{2}) Frame + Frame.
We keep the event branch but replace its input with the frames.
It means the frame/event branch processes the same frames.
\textbf{3}) Frame + Event. 
We take frames and events as input, which is the main method in this paper.
We present the numerical results in Table~\ref{tab:input} and the visual results in Fig. \ref{fig: ablation on input data}.

It can be clearly observed that the setting of frame + event achieves the best result, providing visually pleasing derained images. In Fig. \ref{fig: ablation on input data} (d), (e), we observe that the white pillars are misjudged and removed as rain streaks, which causes the loss of texture details. Meanwhile, when we keep the event branch but replace its input with frames, the additional rainy frames fail to give positive cues for the removal of rain streaks but instead burden the deraining task. In contrast, with the guidance of event data, our proposed method correctly identifies the rain region, which avoids losing the details of the background, especially for the region which is similar to the rain streaks. 
\begin{table}[t!]
\setlength{\tabcolsep}{10pt}
\renewcommand{\arraystretch}{1.2}
\caption{Ablation Results on mapping way and loss function. We predict the clean background layer (background) and predict the negative rain layer (rain) to investigate the influence of different mapping ways. For validating the effectiveness of our proposed multi-scale loss, we additionally devise a single-scale loss that only supervises the last scale output for comparison. The best results are marked in bold.}
\label{tab:abl_mapp_loss}
\resizebox{1\linewidth}{!}{
\begin{tabular}{lccccc}

\toprule[0.8pt]
\multirow{2}{*}{Metrics} & \multicolumn{2}{c}{Mapping Way} && \multicolumn{2}{c}{Loss Function} \\
\cline{2-3} \cline{5-6}
 & background & rain && single-scale & multi-scale \\
\hline
PSNR & 40.59 & \textbf{42.20} && 41.37 & \textbf{42.20} \\
SSIM & 0.9880 & \textbf{0.9903} && 0.9882 & \textbf{0.9903} \\
\bottomrule[0.8pt]

    \end{tabular}
}
\end{table} 
\begin{table}[t!]
\centering
\setlength{\tabcolsep}{10pt}
\renewcommand{\arraystretch}{1.2}
  \caption{Ablation Results on the type of loss functions. We conduct additional experiments with Mean Absolute Error (MAE) loss and Mean Square Error (MSE) loss under the same conditions to validate the effectiveness of negative SSIM loss. The best results are marked in bold.}
  \label{tab:abl_loss}
  \vspace{2mm}
  \resizebox{0.9\linewidth}{!}{
  \begin{tabular}{cccc}
    \toprule[0.8pt]
    Loss Function &    MAE         &    MSE   &  Negative SSIM   \\
    \midrule
    PSNR    &   41.95          &  41.86           &  \textbf{42.20}     \\
    SSIM    &   0.9893         &  0.9891          &  \textbf{0.9903}       \\
  \bottomrule[0.8pt]
\end{tabular}
}
\end{table}

\begin{table}[t!]
\centering
\setlength{\tabcolsep}{10pt}
\renewcommand{\arraystretch}{1.2}
  \caption{Ablation Results on the number of time bins. By uniformly dividing the time interval between adjacent frames into different time bins to investigate the influence of the number of time bins. The best results are marked in bold.}
  \label{tab:num_bins}
  \vspace{2mm}
  \resizebox{0.9\linewidth}{!}{
  \begin{tabular}{ccccc}
    \toprule[0.8pt]
    \#Bin &    5         &    10   &  15 & 20   \\
    \midrule
    PSNR    &   42.08          &  \textbf{42.20}           &  42.15  &  42.20       \\
    SSIM    &   0.9900         &  \textbf{0.9903}          &  0.9902 &  0.9902       \\
  \bottomrule[0.8pt]
\end{tabular}
}
\end{table}
\subsubsection{Influence of Different Modules}

In order to investigate the influences of the components EAMD, REA and ConvLSTM, we ablate each of them to form different variants.
For fair comparisons, we replace the EAMD/REA module with an equal-parameter convolutional block.
For the ablation on ConvLSTM, we only disable the state propagation (by setting the state to zero).
In such a way, the ConvLSTM degrades into a canonical convolutional operation.
We exhibit the quantitative results in Table~\ref{tab:abl_module}.
It can be seen that Model\#D with all components achieves the best result compared with other variants, which verifies the effectiveness of the EAMD module, REA module, and the ConvLSTM block. 

When we substitute the EAMD and REA modules with convolutional blocks of equal parameters, the PSNR decreases to 41.84 dB and 41.69 dB, respectively, from the initial 42.20 dB. This observation confirms that EAMD effectively improves motion-aware regions, while REA enhances the motion characteristics of rain layers. Furthermore, when we disable the state propagation of ConvLSTM, the performance drops to 42.10 dB, underscoring the importance of modeling long-term frame correlations.


\subsubsection{Influence of Mapping Ways}
Instead of using a mapping way to predict the clean background layer, we opt to predict the negative rain layer directly in order to improve the video deraining process. To evaluate the effectiveness of this direct prediction method, we conduct an experiment where we estimate the clean background layer for comparison purposes. The results are presented in Table~\ref{tab:abl_mapp_loss}, demonstrating that directly predicting the residual rain layer is more effective, resulting in a PSNR improvement of over 1 dB. This improvement can be attributed to the simplicity and sparsity of the rain layer, which allows for accurate estimation when compared to the background layer.

\subsubsection{Influence of Loss Functions}
To enhance the training of our network, we establish a multi-scale loss as detailed in Section \ref{sec:loss}. To assess its efficacy, we create a single-scale loss, which solely supervises the final scale output. Furthermore, we perform experiments using MAE and MSE loss functions under identical conditions to validate the potency of the negative SSIM loss. The numerical outcomes in Table~\ref{tab:abl_mapp_loss} and Table~\ref{tab:abl_loss} conclusively demonstrate that our multi-scale negative SSIM loss significantly enhances our network's performance, affirming its effectiveness.

\subsubsection{Influence of the number of Time Bins}
We examine how the number of bins in the event voxel grid affects our investigation. We evenly distribute events into varying bin counts for our experiments. In Table \ref{tab:num_bins}, we find that our method is not significantly affected by the number of time bins. We identify a 10-bin voxel grid as the optimal choice and apply it in all previous experiments.

\section{Conclusion}
In this paper, we present a learning-based framework for addressing the video deraining task using an event camera. Our approach comprises two key components: an event-aware motion detection module and a pyramidal adaptive selection module. These modules are designed to effectively enhance motion-aware regions and extract rain layers. Furthermore, we have curated a real-world dataset specifically for event-based video deraining. We provide quantitative and qualitative evidence showcasing the superiority of our method compared to state-of-the-art techniques, across both synthetic and real-world datasets. We anticipate that the concepts we propose can find application in restoring clear videos in adverse weather conditions like snow, hail, and sandstorms, and we plan to explore these possibilities in our future work.

\bibliographystyle{IEEEtran}
\bibliography{wj}

\end{document}